\documentclass[10pt,twocolumn,letterpaper]{article}

 \usepackage[pagenumbers]{iccv} 

\definecolor{iccvblue}{rgb}{0.21,0.49,0.74}
\usepackage[pagebackref,breaklinks,colorlinks,allcolors=iccvblue]{hyperref}

\usepackage{mathrsfs}
\usepackage{amssymb}
\usepackage{amsmath}
\usepackage{dsfont}
\usepackage{tablefootnote}
\usepackage{float}
\usepackage{colortbl}
\usepackage{algorithm}
\usepackage[noend]{algpseudocode}
\usepackage{multirow} 
\usepackage{xcolor}

\newcommand{\group}{\mathcal{G}}

\usepackage{mathtools}
\usepackage{amsthm}

\usepackage[capitalize,noabbrev]{cleveref}
\usepackage{lipsum}

\def\paperID{9534} 
\def\confName{ICCV}
\def\confYear{2025}

\title{Post-Training Quantization for Diffusion Transformer \\via Hierarchical Timestep Grouping}

\author{
Ning Ding$^{1*}$, Jing Han$^{2}$\thanks{Equal Contribution. $\dagger$Corresponding Author.}\,\,, Yuchuan Tian$^{1}$, Chao Xu$^{1}$, Kai Han$^{3}$, Yehui Tang$^{3\dagger}$ \\
	\small$^1$ State Key Lab of General AI, School of Intelligence Science and Technology, Peking University \\
	 \small$^2$ School of Artificial Intelligence, Beijing University of Posts and Telecommunications \qquad \small$^3$ Huawei Noah's Ark Lab.\\
	\small\texttt{dingning@stu.pku.edu.cn \quad yehui.tang@huawei.com}\\
}

\begin{document}
\maketitle

\begin{abstract}

	Diffusion Transformer (DiT) has now become the preferred choice for building image 
	generation models due to its great generation capability. Unlike previous convolution-based UNet models, DiT is purely composed of a stack of transformer blocks, which renders DiT excellent in scalability like large language models. 
	However, the growing model size and multi-step sampling paradigm bring about considerable pressure on deployment and inference.
	In this work, we propose a post-training quantization framework tailored for Diffusion Transforms to tackle these challenges. 
	We firstly locate that the quantization difficulty of DiT mainly originates from the time-dependent channel-specific outliers.
	We propose a timestep-aware shift-and-scale strategy to smooth the activation distribution to reduce the quantization error.
	Secondly, based on the observation that activations of adjacent timesteps have similar distributions, we utilize a hierarchical clustering scheme to divide the denoising timesteps into multiple groups. 
	We further design a re-parameterization scheme which absorbs the quantization parameters into nearby module to avoid redundant computations.
	Comprehensive experiments demonstrate that out PTQ method successfully quantize the Diffusion Transformer into 8-bit weight and 8-bit activation (W8A8) with state-of-the-art FiD score. And our method can further quantize DiT model into 4-bit weight and 8-bit activation (W4A8) without sacrificing generation quality.

\end{abstract}

\section{Introduction}
\label{sec:intro}

Diffusion models have achieved state-of-the-art performance in various generation tasks, including text-to-image generation, image-conditioned style transformation, \textit{etc}. Traditional diffusion models are build upon U-Net architectures~\cite{song2020denoising,ho2020denoising}, relying mainly on hierarchical convolutional layers to capture local features and contextual information in images. However, such convolution-base network is faced with scalability issue and its capability will quickly saturate even with increased training data. As visual generation tasks getting more diversified, U-Net architecture shows limitations in handling more complex situations such as video generation.
Consequently, recent researches have focused on diffusion models based on self-attention mechanism and transformer style network. This type of diffusion model, known as Diffusion Transformer~\cite{peebles2023scalable} (DiT), can better capture global information in images and can handle larger image resolutions. Compared to traditional convolutional architecture, DiTs also enjoy great scalability just like large language models, which can fit larger datasets and model sizes. Famous commercial frameworks like Stable Diffusion 3\cite{esser2024scaling} and Sora\cite{liu2024sora} have scaled DiT architectures up to billions of parameters and demonstrated significant image- and video-generation ability.

Despite the impressive performance improvement achieved by DiTs, Transformer-based models have some inherent disadvantages when it comes to deployment and inference. Firstly, DiT models typically contain a large number of parameters, resulting in higher storage requirements. Secondly, the computational complexity of self-attention mechanism grows quadratically with the increase of image resolutions. Therefore, the peak memory usage and the time-consumption of the inference process are both more demanding than convolutional models. Thirdly. diffusion models need to iteratively run forward-pass for multiple times during inference. These issues make it difficult for Diffusion Transformers to meet the real-time and efficiency requirements in practical applications, especially posing challenges for the deployment on resource-constrained devices such as cellphones.

Post-training quantization (PTQ) is an effective technique to combat the above difficulties. Without needing any training data, PTQ method compresses the model size by converting the 32-bit floating point weights into 8-bit or 4-bit integers, which directly minimizes the storage requirement. More importantly, PTQ method can also convert the activations (\textit{e.g.} the input of a linear layer) from 32-bit float numbers to 8-bit integers. As a result, the matrix multiplications of both attention module and linear layer take place in the low-precision integer field, thus accelerating the inference process and reducing the memory footprint.
Nevertheless, quantizing the Diffusion Transformer (DiT) presents significant challenges. Through a detailed analysis of the activation distribution in the DiT model, we identify two key issues lies ahead:  (1) Extreme values with large magnitude exist in activations and the presence of outliers is associated with specific activation channels. Given the imbalanced and asymmetric distribution, these outliers cause huge quantization error to the majority of activations with small values. (2) Even if we only observe the outliers within the same channel, their distribution varies across different denoising timesteps. This phenomenon will render the calibrated activation quantizer invalid in static post-training quantization scenario.

In this paper, we propose the Hierarchical Timestep Grouping (HTG) framework, a novel post-training quantization method designed for Diffusion Transformers, to tackle the above challenges.
One the one hand, we propose the temporally-grouped channel-wise shifting to dynamically adjust and re-distribute the asymmetric distribution of outliers in different channels, which reduces the quantization error for non-outlier activations. We then aggregate neighboring timesteps into multiple temporal groups using hierarchical clustering, based on the similarity of the skewness of outlier channels across timesteps.
On the other hand, we propose the temporally-aggregated channel-wise scaling to transfer the quantization difficulty from multi-step's activations to the weight matrix of the linear layer, which further suppresses the magnitude of outliers throughout the denoising process. We derive a scaling vector that can generalize to every timestep based on history outlier information to facilitate the uniform quantization of the static activation quantizer.
Finally, we design a re-parameterization scheme to merge the shifting and scaling vectors into the preceding AdaLN module, avoiding the introduction of additional unnecessary computations.

The main contributions of this paper are as follows:
\begin{itemize}
	
	\item We propose a static post-training quantization (PTQ) framework termed Hierarchical Timestep Grouping (HTG), to alleviating the quantization difficulties for DiT models by smoothing the temporally-varied activations. Our method runs out-of-the-box without the need for re-training or fine-tuning. 
	\item We achieve the state-of-the-art performance for quantized DiT models in both 8-bit weight 8-bit activation (W8A8) and 4-bit weight 8-bit activation (W4A8) settings.
\end{itemize}

\section{Related Works}

\textbf{Diffusion models}~\cite{sohl2015deep, ho2020denoising, song2020score} gradually add noise to the data and then reverse this process to generate data from the noise. The forward noise-injection process is modeled as a Markov chain:
\begin{equation} \notag
	q(x_{1:T}|x_0) = \prod_{t=1}^T q(x_t|x_{t-1}).
\end{equation}
In this formulation, $x_0$ represents the original data, and $q(x_t|x_{t-1}) = \mathcal{N}(x_t|\sqrt{\alpha_t} x_{t-1}, \beta_t \mathbb{I})$ defines the transition probability, where $\alpha_t$ and $\beta_t$ are the elements of the noise schedule, constrained by $\alpha_t + \beta_t = 1$. To reverse this process, a Gaussian model $ p(x_{t-1}|x_t) = \mathcal{N}(x_{t-1}|\mu_t(x_t), \sigma_t^2 \mathbb{I})$ is employed to approximate the real reverse transition $q(x_{t-1}|x_t)$,
where an optimal estimation~\cite{bao2022analytic} for this reverse process is given by
\begin{equation} \notag
	\mu_t^*(x_t) = \frac{1}{\sqrt{\alpha_t}} \left(x_t - \frac{\beta_t}{\sqrt{1-\overline{\alpha}_t}} \mathbb{E}[\epsilon|x_t]\right).
\end{equation}
Here, $\overline{\alpha}_t = \prod_{i=1}^t \alpha_i$, and $\epsilon$ represents the noise that has been added to $x_t$. The learning task is thus equivalent to predicting this noise. Specifically, a noise prediction network $\epsilon_\theta(x_t, t)$ is trained to estimate $\mathbb{E}[\epsilon|x_t]$ by minimizing the following objective function:
\begin{equation} \notag
	\min_{\theta} \mathbb{E}_{t, x_0, \epsilon} \| \epsilon - \epsilon_\theta(x_t, t) \|_2^2.
\end{equation}
where $t$ is sampled uniformly between 1 and $T$. To extend this to conditional diffusion models, such as class-conditional~\cite{dhariwal2021diffusion} or text-to-image~\cite{ramesh2022hierarchical} generation, additional condition information is incorporated into the noise prediction network, modifying the objective as 
\begin{equation} \notag
	\min_{\theta} \mathbb{E}_{t, x_0, c, \epsilon} \| \epsilon - \epsilon_\theta(x_t, t, c) \|_2^2,
\end{equation}
where $c$, usually modeled by continuous embeddings, represents the prior condition. The Diffusion Transformer~\cite{peebles2023scalable} is used as $\epsilon_\theta$ in this literature.

\textbf{Post-Training Quantization} is a wildly used technique that compresses the model size by converting the full-precision weights into low precision integers.
More importantly, PTQ can reduce the GPU-memory footprint and accelerate the inference speed from the hardware level. Given a floating-point tensor $\bold X$, the quantization function $Q(\cdot)$ converts $\bold X$ to $b$-bit integers by
\begin{equation}
	\label{eq:ptq}
	Q(x) = \text{clamp}(\lfloor \frac{\bold X}{\Delta} \rfloor) + \lambda , 0, 2^b-1),
\end{equation}
where $\lfloor\cdot\rfloor$ is the floor function, $\Delta$ is a floating-point scalar to control the width of quantization level, $\lambda$ is an integer to control the symmetry of quantized values, and clamp($\cdot$) function truncates the input to guarantee the range of $Q(x)$ falling into $[0, 2^b-1]$.
The post-training quantization methods have been extensively explored for various types of deep models, such as CNN~\cite{liu2023pd,wei2022qdrop,lee2023flexround,nagel2020up}, ViT~\cite{li2023vit,li2023repq,yuan2022ptq4vit,yang2024mgrq}, \textit{etc}.
With the rapid development of large language models, the research~\cite{pmlr-v202-xiao23c,wei2023outlier,wei2022outlier,shao2023omniquant,lin2024awq} on PTQ methods for the transformer architecture draws much attention, since they address practical deployment challenges by reducing the memory consumption and inference latency.
Recently, some works~\cite{wu2024ptq4dit,chen2024q} proposed different solutions to quantize the Diffusion Transformers. PTQ4DiT~\cite{wu2024ptq4dit} focuses on removing the activation outlier by scaling their magnitude while neglecting the asymmetric distribution of outlier channels across different timesteps. Q-DiT~\cite{chen2024q} utilizes the dynamic activation quantizer technique for DiT models, which might slows down the inference speed.
In this work, we focus on \textbf{static} post-training quantization where the activation quantizer uses \textbf{per-tensor} uniform quantization.

\begin{figure*}[t]
	\centering
	\begin{subfigure}[b]{0.73\textwidth}
		\includegraphics[width=0.99\textwidth]{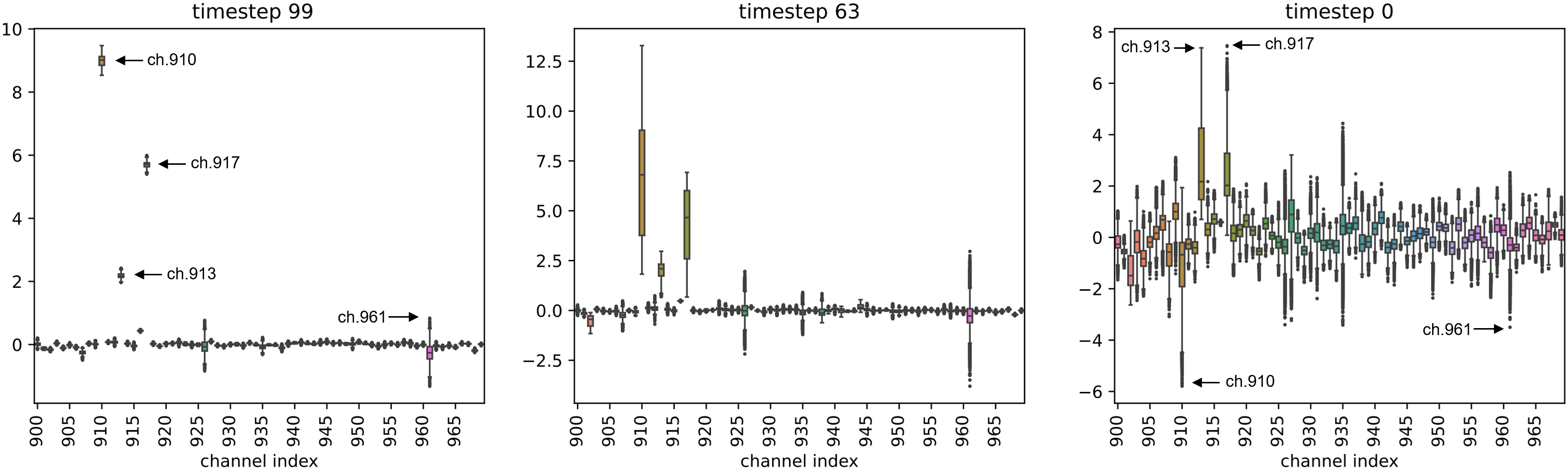}
		\caption{}
		\label{fig:outlier}
	\end{subfigure}
	\vline
	\begin{subfigure}{0.23\textwidth}
		\includegraphics[width=0.99\textwidth]{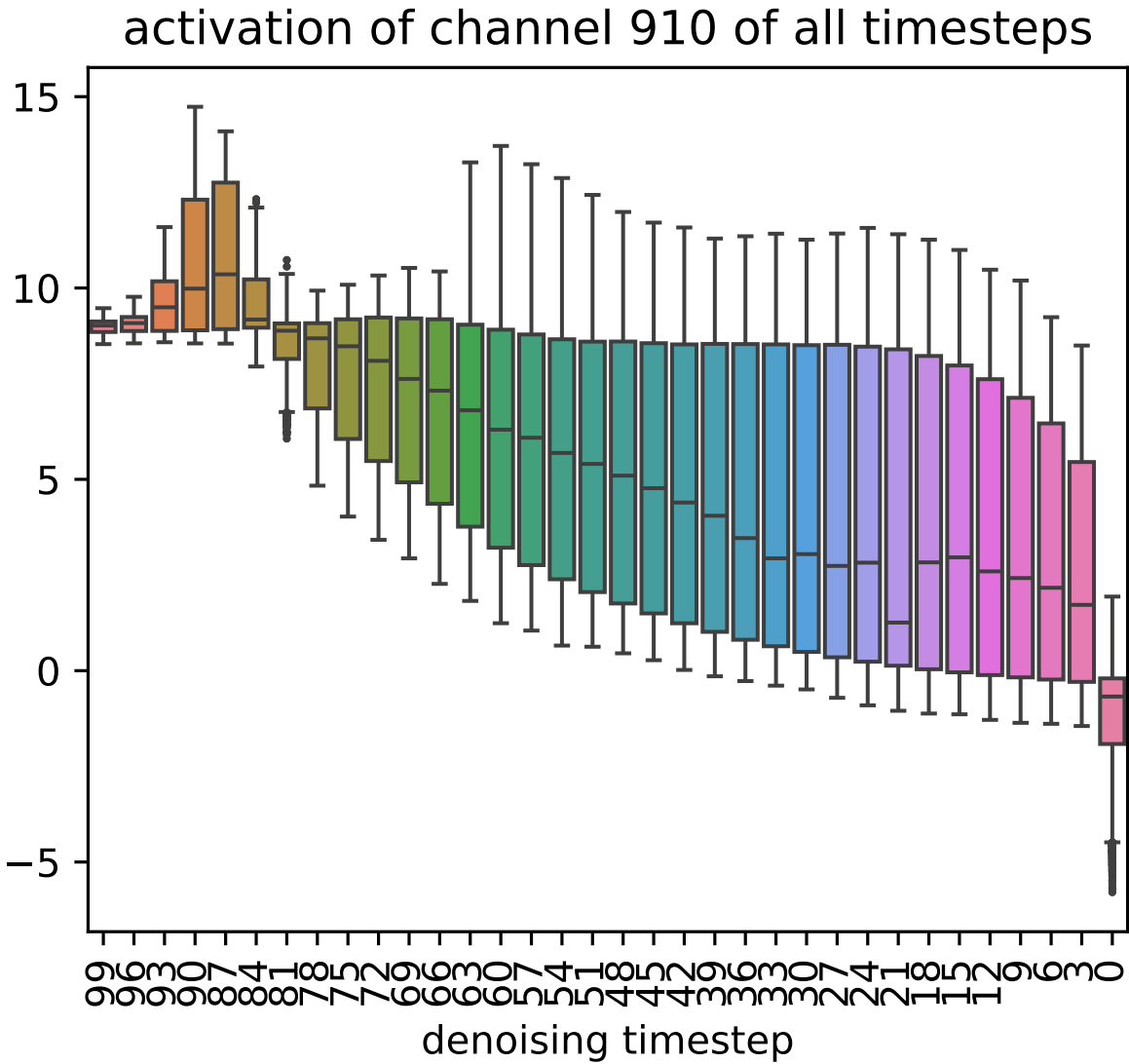}
		\caption{}
		\label{fig:all_timestep}
	\end{subfigure}
	\vspace{-2mm}\caption{(a) Per-channel input activation of blocks[27].mlp.fc1 at step 99, step 63 and step 0. Channels with indexes between 900-970 are displayed. (b) Activations of channel No.910 across different timesteps of 100-step DDPM scheduler.}
\end{figure*}

\vspace{-2mm}
\section{Reviewing Quantization Difficulty in DiT}
\vspace{-1mm}
As for traditional diffusion models, there are a large number of convolutional and down-sampling operations in the U-Net architecture, which can smooth the activations (feature maps), thus helping to mitigate the quantization difficulties.
The Diffusion Transformer (DiT) is constructed upon a pile of stacked transformer blocks which are comprised of self-attention operations and MLP modules. The activation distributions of DiT are nothing like convolutional models. Combined with multi-timestep iterative inference, these characteristics make DiT models more difficult to quantize.
In this section, we analyse the main challenges posed by the post-training quantization of DiTs.

\subsection{Channel-dependent activation outliers}
\vspace{-1mm}
\label{difficulty1}
The biggest quantization challenge for transformer architecture is the outlier phenomenon that appears in some specific channels of the input activation of linear layers. The outlier refers to extreme features whose value range is significantly deviated from the average level. 
As shown in \cref{fig:outlier}, we visualize the input activations at multiple denoising timesteps for the 27th block's \texttt{mlp.fc1} layer of the DiT-XL/2 model with a 100-step DDPM~\cite{ho2020denoising} scheduler.
The usual practice of uniform quantization is to truncate these extreme outliers in order to provide sufficient quantization resolution for the majority of activations distributed near 0, since directly quantizing these extreme values would waste most of the quantization levels thus causing huge quantization error. This is unacceptable for DiT models where the error accumulates exponentially throughout the iterative diffusion process.
However, outliers often carry important information and truncating them directly would lead to distorted generation result.

\subsection{Timestep-dependent outlier distributions}

Apart from the outlier problem mentioned above, another difficulty in quantizing DiT comes from the multi-timestep generating paradigm of the diffusion model, which is intractable for static PTQ method using one set of pre-computed quantization parameter for each layer's activation.
\cref{fig:all_timestep} demonstrates the activation distribution of 910th channel for the 27th block's \texttt{mlp.fc1} layer of the DiT-XL/2 model. The activation median is around 10 at the early denoising stage, and decreases to $\sim$3 as timestep getting smaller. The extreme value of this channel suddenly becomes -6 at the final denoising step while the extreme values of other timestep all remain positive (mostly $>$10). This example indicates that the distribution of outlier at one specific channel varies drastically across different timesteps of the denoising process.
In this case, the $\delta$ and $\sigma$ in \cref{eq:ptq} obtained from early-stage calibration data will be invalid for the activations of later-stage inference.

\section{Hierarchical Timestep Grouping}

In this section, we elaborate the proposed Hierarchical Timestep Grouping (HTG) framework, a post-training quantization method specially tailor for Diffusion Transformers. 
We use bold lowercase letters (\eg~$\mathbf{x}$) to denote vectors, bold uppercase letters to denote matrices (\eg~$\mathbf{W}$).
We use the notation $[\cdot]_i$ to represent the $i$th column of a matrix, and also use $[\cdot]_i$ to represent the $i$-th entry of a vector. The $\text{diag}(\cdot)$ function turns a vector in to a diagonal matrix where all elements except the diagonals are zeros.

\begin{figure}[t]
	\centering
	\includegraphics[width=0.49\textwidth]{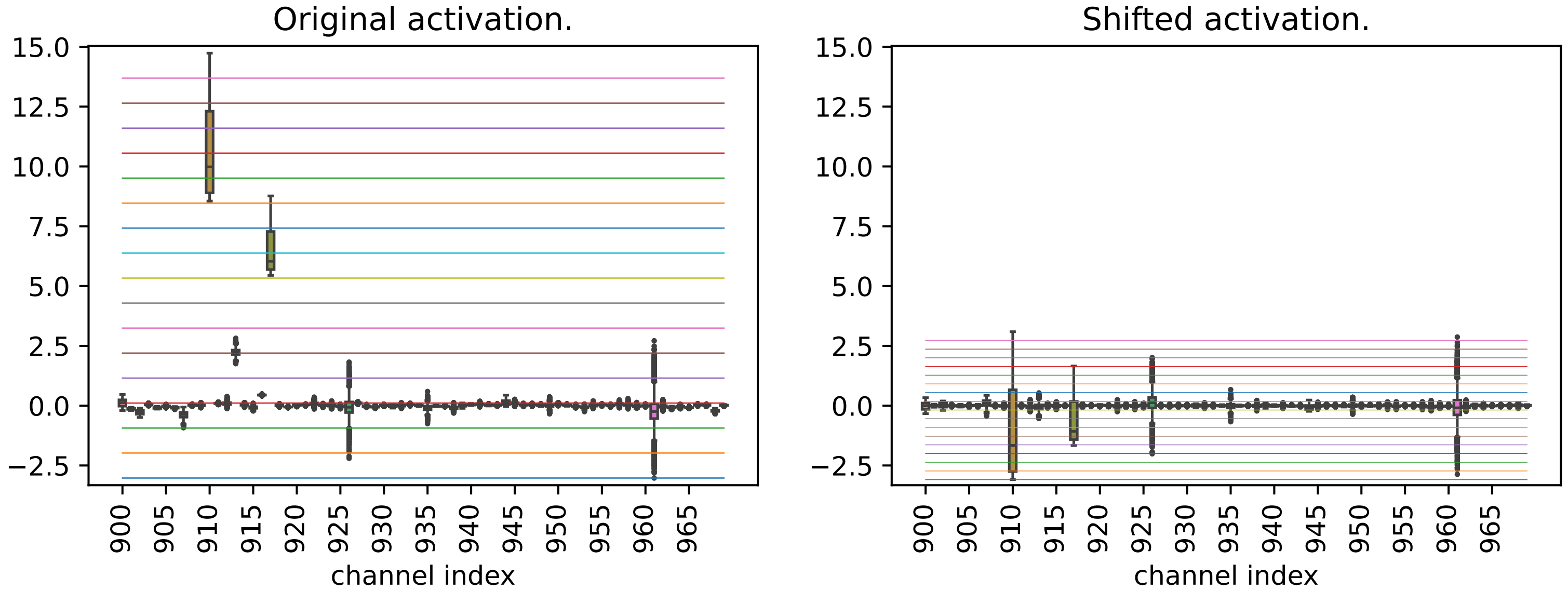}
	
	\vspace{-2mm}\caption{Original activation (left) and shifted activation (right). Activations are taken from blocks[27].mlp.fc1 at step 90.}
	\label{fig:single_shift}
\end{figure}

\subsection{Temporally-grouped Channel-wise Shifting}
\label{Channel-wise-shifting}
For a linear layer parameterized by $\bold W\in\mathbb R^{C_{in}\times C_{out}}$, it takes $\bold X\in\mathbb R^{n\times C_{in}}$ as input and performs $\bold Y=\bold X \bold W + \bold b$, where $n$ is the length of the token sequence, $C_{in}$ is the number of input channels, $C_{out}$ is the number of output channels, and $\bold b\in\mathbb{R}^{C_{out}}$ is the bias term, respectively.
As we discussed in \cref{difficulty1}, the outlier issue is closely related to specific channels of $\bold X$. \cref{fig:single_shift} showcases a 4-bit quantization scenario where all activations are uniformly partitioned into 16 bins (quantization levels). However, the extreme values occupy absolutely the most of the range of positive side. The serious asymmetry in the outlier distribution leads to the vast majority of common activations being squeezed into only one or two bins, which corresponds to huge quantization error $||\bold X_t - Q(\bold X_t)||_2$.

To handle this issue, we first use a channel-wise shifting vector to adjust the per-channel distributions and make the asymmetric outliers zero-centered to ease the quantization difficulty. For any given denoising timestep $t$, activation shifting is expressed by:
\begin{equation}\label{first_shift}
\begin{aligned}
	&\widetilde{\bold X}_t = \bold X_t - \bold z_t, ~ t\in\{T,T-1,\cdots,2,1\}, \\
	\text{\small where}~&\bold z_t\in\mathbb R^{C_{in}}, ~[\bold z_t]_i = \frac{\max([\bold X_t]_i) + \min([\bold X_t]_i)}{2}.
\end{aligned}
\end{equation}
As visualized on the right side of \cref{fig:single_shift}, activations of non-outlier channels can enjoy more quantization bins after the shifting operation, and therefore the quantization error $||\widetilde{\bold X}_t - Q(\widetilde{\bold X}_t)||_2$ is reduced.

However, too much storage space would be consumed if we were to store one shifting vector for each denoising timestep. Taking a 100-step diffusion process as an example, a single linear layer will need to store an extra matrix of the size $\mathbb R^{C_{in}\times 100}$ to deal with the temporally varying activations. Therefore, we need to find a way to compress these shifting vectors for minimum storage overhead. Recent works~\cite{park2023denoising,yuan2024ditfastattn,teng2024accelerating} have shown that the distributions of activations in the same layer are similar for neighboring timesteps during the diffusion process. The visualization shown in \cref{fig:all_timestep} is in agreement with this discovery, where the ranges of values for the activations in the same channel changes smoothly during continuously adjacent timesteps. Leveraging this property, we divide all timesteps into several groups according to the distributions of outliers.

\begin{algorithm}[t]
\caption{Constrained hierarchical clustering}
\label{algo_cluster}
\footnotesize
\begin{algorithmic}[1]
	\Require $T$ shifting vectors$\{\bold z_t\}_{t=1}^T$, target number of groups $G$.
	\State Treating every $\bold{z}_t$ as a cluster to construct $T$ initial groups;
	\State Compute the paired distance $Dist_{i,j}(\text{Group}_i, \text{Group}_j)$ between all adjacent groups where $|i-j|=1$;
	\State Merge the $i_{th}$ and $j_{th}$ group with minimum $Dist_{i,j}$ into one group;
	\State Repeat (2) and (3) until the current number of groups is reduced to $G$.
	\State \Return Partition boundaries $\{\tau_1, \tau_2, \cdots, \tau_{G-1} \}$.
\end{algorithmic}
\end{algorithm}

Given that the shifting vector $\bold z_t$ reflects the skewness of each channel's activation at timestep $t$,  we can divide the denoising steps into $G$ groups($G$ is much smaller than $T$) based on the similarities in shifting vectors. This can be formulated as a constrained hierarchical clustering~\cite{gower1969minimum, jarman2020hierarchical} on $\bold z_t$ with $t\in\{T,T-1,\cdots,2,1\}$, where the constraint is that the timesteps belonging to the same group are continuous. This problem is also equivalent to finding the $G-1$ partition boundaries that divides $\bold z_t$ into $G$ groups.
Supposing $\group(\cdot)$ is a function to map a timestep to its group index $g=\group(t)\in\{1,2,\cdots,G\}$, and $\tau_i$ is the partition boundaries where $i\in\{1,2,\cdots,g-1\}$, we need to solve the optimization problem:
\begin{equation}
\begin{aligned}
	\label{hier_cluster}
	& \mathop{\arg\min}\limits_{\tau_1, \tau_2,\cdots,\tau_{G-1}} \sum_{g=1}^{G} \sum_{t=1}^{T} \mathds{1}_{\group(t)=g}|| \bold{z}_t - \mu_g||_2 \, , \\
	&\textbf{s.t.}
	\left\{
	\begin{aligned}
		& \, \tau_0=0, \, \tau_{G}=T  \, , \\
		& \,  0< \tau_1< \tau_2<\cdots<\tau_{G-1}<T \,, \\
		&  \, \tau_{g-1} < t \le \tau_{g}\,\, \text{for}\,\, \forall \group(t)=g \,. 
	\end{aligned}
	\right.
\end{aligned}
\end{equation}
where $\mu_g$ is the cluster centriod of the $g_{\text{th}}$ group, and $\mathds{1}_{condition}$ is the indicator function which equals to 1 when the condition is met or otherwise 0.
\cref{hier_cluster} can be solved by performing \cref{algo_cluster}.
The computation of the paired distance could be implemented by average-linkage method~\cite{seifoddini1989single}, centroid-linkage method~\cite{jarman2020hierarchical}, Ward method~\cite{murtagh2014ward} and \textit{etc}.
After clustering, we further compress the shifting vectors by taking the average of $\bold{z}_t$ within the same temporal group:
\begin{equation}\label{computer_group_shift}
	\overline{\bold z}_{g} = \sum_{t=1}^{T} \frac{\mathds 1_{\group(t)=g}~\bold{z}_t}{ \sum_{t=1}^{T}\mathds 1_{\group(t)=g}},
\end{equation}
and therefore, the activation $\bold X_t$ will use $\overline{\bold z}_{\group(t)}$ for distribution re-adjustment.
So far, we have successfully compressed the $T$ shifting vectors into $G$ vectors. Not only does this greatly reduce the storage overhead, but we also discover via ablation study that this operation hardly affects the quantization performance compared to preserving all $\bold z_t$.

\cref{first_shift} now becomes$\widetilde{\bold X}_t = \bold X_t - \overline{\bold z}_{\group(t)}$. The subsequent linear layer which takes $\bold X_t$ as input now computes

\begin{equation}
	\begin{aligned}
		\bold Y &=  \bold X_t\bold W  + \bold b \\ 
		&= (\bold X_t - \overline{\bold z}_{\group(t)})\bold W + (\bold b + \overline{\bold z}_{\group(t)}\bold W ) \\ 
		&=\widetilde{\bold X}_t \bold W + \widehat{\bold b}_{\group(t)}  \label{shift_linear}
	\end{aligned}
\end{equation}
to keep an equivalent transformation. \cref{shift_linear} suggests that the linear layer stores $G$ biases to compensate for the corresponding channel-wise outlier shifting operations, where the new bias term that corresponds to the $g_{\text{th}}$ temporal group is given by $\widehat{\bold b}_{g}=\bold b + \overline{\bold z}_{g}\bold W$. 

\begin{figure}[t]
	\centering
	\includegraphics[width=0.49\textwidth]{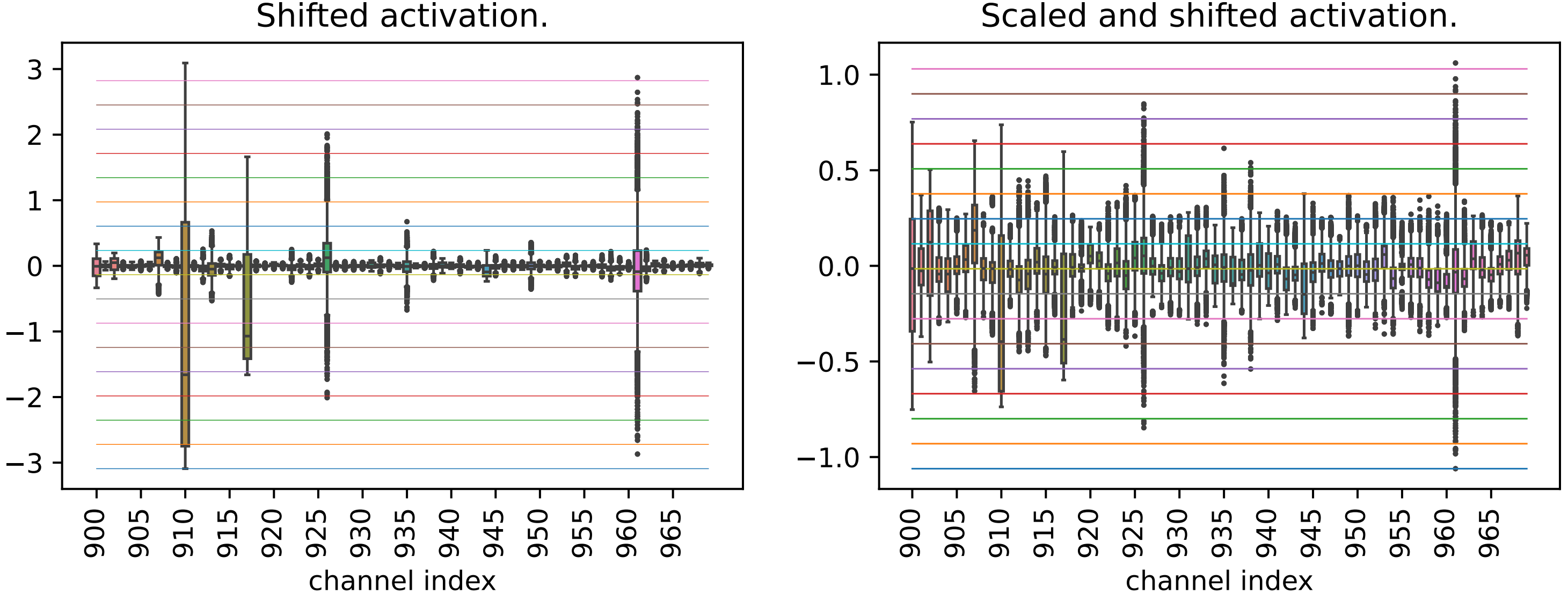}
	\vspace{-5mm}\caption{Scaling the shifted activation leads to further less quantization error for non-outliers. Activations are taken from blocks[27].mlp.fc1 at step 90.}
	\label{fig:scale_shift}
\end{figure}

\subsection{Temporally-aggregated Channel-wise Scaling}
\label{Channel-wise-scaling}
Given a channel-wise shifted activation, the range of floating-point value may still be dominated by the magnitude of outlier channel. Existing PTQ algorithms (\eg~SmoothQuant~\cite{pmlr-v202-xiao23c}) proposed to mitigates this imbalance by simultaneously scaling the activation and weight matrix in the opposite direction, thus transferring the quantization difficulty from activations to weights.
Considering the shifted activation derived in the previous subsection, each channel's values of $\widetilde{\bold X}_t$ are symmetric about zero after shifting. Therefore we can define the per-channel smoothing vector $\bold s_t\in\mathbb{R}^{C_{in}}$, at timestep $t$, as
\begin{equation} \notag
[\bold s_t]_i =\sqrt{ \frac{\max([\widetilde{\bold X}_t]_i)}{\max([\bold W^\top]_i)} },
\end{equation}
and the linear layer is reformulated to
\begin{equation}
\begin{aligned}
	\bold Y &=  \widetilde{\bold X}_t \bold W  + \bold b  \\ 
	&=  (\widetilde{\bold X}_t \text{diag}(\bold s_t)^{-1}) (\text{diag}(\bold s_t) \bold W)  + \bold b  \\
	&= \hat{\bold X}_t \hat{\bold W_t} + \bold b  \label{scale_linear}
\end{aligned} 
\end{equation}
to keep mathematical equivalency.
Although \cref{scale_linear} manages to smooth the outliers by reversely amplifying the magnitude of the corresponding channel of the weight matrix, yet this naive method introduces an individual weight matrix $\hat{\bold W_t}=\text{diag}(\bold s_t) \bold W$ for each timestep $t$, which is a prohibitively huge cost to save multiple checkpoints in PTQ scenario. 

In order to maintain the global consistency of the weight matrix, we propose to aggregate the magnitude information for each channel across all timesteps into one single scaling vector $\overline{\bold s}$ to represent the overall statistics of outliers during the whole denoising process.
Specifically, we resort to the exponential moving average. During the inference process, we set up a buffer $\bold m$ to keep a history record of the activation outlier of each channel. The $\bold m$ at step $t$ is given by:
\begin{equation}\label{eq:ema}
	[\bold m_t]_i = \alpha \bold m_{t+1} + (1-\alpha) \max([\widetilde{\bold X}_t]_i), t\in\{T-1,\cdots,2,1\}, 
\end{equation}
where $[\bold m_T]_i = \max([\widetilde{\bold X}_T]_i)$, $\alpha$ is empirically set to 0.99.
As $t$ decreases from $T$ to 1, the scaling vector $\overline{\bold s}$ is derived using the $\bold m$ of the final step:
\begin{equation}
	[\overline{\bold s}]_i =\sqrt{ \frac{[\bold m_1]_i}{\max([\bold W^\top]_i)} }. \notag
\end{equation}
This approach statistically finds a suitable value that represents the magnitude of channel outlier across all timesteps.
Combined with the temporally-grouped per-channel shifting operation in \cref{shift_linear}, the complete Hierarchical Timestep Grouping is formulated as 
\begin{equation}
\label{scale_and_shift_linear}
\begin{aligned}
	\bold Y &=  \bold X_t\bold W  + \bold b \\ 
	&= (\bold X_t - \overline{\bold z}_{\group(t)})\text{diag}(\overline{\bold s})^{-1}\text{diag}(\overline{\bold s})\bold W + (\bold b + \overline{\bold z}_{\group(t)}\bold W ) \\ 
	&=\widehat{\bold X}_t \widehat{\bold W} + \widehat{\bold b}_{\group(t)},
\end{aligned}
\end{equation}
where $\widehat{\bold W} = \text{diag}(\overline{\bold s})\bold W$ is the re-scaled weight matrix to be quantized and
\begin{equation}
	\label{x_to_be_merged}
	\widehat{\bold X}_t=(\bold X_t - \overline{\bold z}_{\group(t)})\text{diag}(\overline{\bold s})^{-1}
\end{equation}
is the smoothed activation.

\cref{fig:scale_shift} demonstrates a 4-bit quantization scenario, where the activations of non-outlier channel are allocated with more quantization levels after scaling with \label{scale_and_shift_linear}.

\begin{figure}[t]
	\centering
	\includegraphics[width=0.49\textwidth]{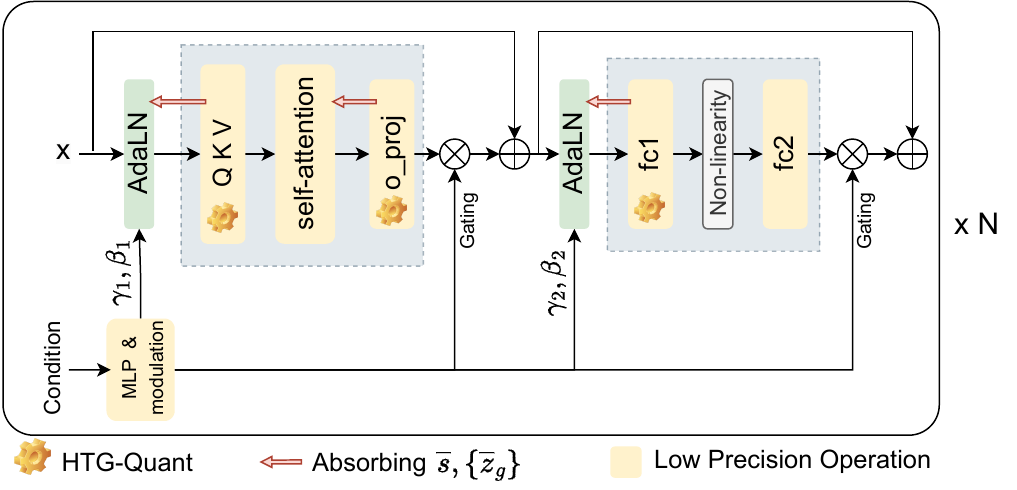}
	\vspace{-6mm}\caption{Schematic diagram of the HTG framework for DiT.}
	\label{fig:quant_diagram}
\end{figure}

\subsection{Re-parameterization for Faster Inference}
Before quantizing the DiT model, we firstly calculate the $\overline{\bold s}$ and $\{\overline{\bold z}_g\}_{g=1}^{G}$ for each linear layer by using some calibration data generated for each timestep.
To avoid introducing any additional computations into the inference process, we cannot perform online shifting and scaling operation mentioned above to the input activations. We design a re-parameterization method~\cite{Ding_2021_CVPR,wang2024repvit,wu2024ptq4dit} for DiT architecture which allows the shifting and scaling vectors to be absorbed by the AdaLN~\cite{peebles2023scalable} module preceding the linear layer.

Denoted by $\bold Z$ the hidden state of the token sequence, AdaLN performs
\begin{equation}\label{adaln}
	\bold X=\text{AdaLN}(\bold Z) = \text{LayerNorm}(\bold Z)(1+\boldsymbol{\gamma})+ \boldsymbol{\beta},
\end{equation}
in which $\boldsymbol{\gamma},\boldsymbol{\beta}\in\mathbb{R}^{C_{in}}$ are learned parameters. By decomposing \cref{x_to_be_merged} and merging $\overline{\bold s}$ and $\{\overline{\bold z}_g\}_{g=1}^{G}$ into \cref{adaln}, we have the re-parameterized $\widehat{\text{AdaLN}}$ given by:

\begin{align} \notag
	\widehat{\bold X}_t=\widehat{\text{AdaLN}}& (\bold Z_t) = \text{LayerNorm}(\bold Z_t)(\frac{1}{\overline{\bold s}}+\widehat{\boldsymbol{\gamma}}) + \widehat{\boldsymbol{\beta}}_{\group(t)}   , \\
	\text{where}\quad \widehat{\boldsymbol{\gamma}}& = \frac{\boldsymbol{\gamma}}{\overline{\bold s}}, \quad
	\widehat{\boldsymbol{\beta}}_{\group(t)} = \frac{\boldsymbol{\beta}-\overline{\bold z}_{\group(t)}}{\overline{\bold s}}. \label{reparam_adaln}
\end{align}
After the re-parameterization, we convert $\bold W$ into $\widehat{\bold W}$, and then apply the uniform quantization to weight matrices of the linear layers in the DiT model. 
It should be noticed that the proposed Hierarchical Timestep Grouping (HTG) method is not only performed at \texttt{attn.qkv} and \texttt{mlp.fc1} where AdaLN precede them. The HTG is also performed at \texttt{attn.o$\_$proj} since the shifting and scaling vectors can be merge into the de-quant operation of the preceding self-attention module.
\cref{fig:quant_diagram} showcases our HTG quantization scheme for the DiT model.
The $\widehat{\text{AdaLN}}$ and succeeding linear layer merely need to occasionally switch to the corresponding bias term $\widehat{\boldsymbol{\beta}}_g$ and $\widehat{\bold b}_g$, respectively, at some certain timesteps when a new temporal group begins. Therefore, our framework does not introduce additional bit-operation. \cref{algo} elucidates the complete procedure for quantizing the DiT model using the proposed Hierarchical Timestep Grouping method.

\begin{algorithm}[t]
	\caption{Hierarchical Timestep Grouping}\label{algo}
	\footnotesize
	\begin{algorithmic}[1]
		\Require \texttt{attn.qkv}, \texttt{mlp.fc1} and \texttt{attn.o$\_$proj} layers in pretrained DiT model; Activation sequence $\{\bold X_t\}_{t=1}^T$ generated by calibration dataset.
		
		\State Compute the channel-wise shifting vector $\{\bold z_t\}_{t=1}^T$; \Comment{\cref{first_shift}}
		\State Partition all timesteps into $G$ temporal groups; \Comment{\cref{algo_cluster}}
		\State Compute grouped channel-wise shifting $\{ \overline{\bold z}_g\}_{g=1}^G$ and corresponding bias terms $\{\widehat{\bold b}_g\}_{g=1}^G$; \Comment{\cref{computer_group_shift} \& \cref{shift_linear}}
		\State Compute the channel-wise scaling vector $\overline{\bold s}$;
		\Comment{\cref{eq:ema}}
		\State Rescaled the weight matrix $\widehat{\bold W} = \text{diag}(\overline{\bold s})\bold W$;
		\State Absorbing $\{ \overline{\bold z}_g\}_{g=1}^G$ and $\overline{\bold s}$ into preceding AdaLN module; \Comment{\cref{reparam_adaln}}
		\State Quantize $\widehat{\bold W}$ and $\widehat{\bold X}$ with uniform quantizer.
		\State Quantize $\bold W$ and $\bold X$ in the remaining linear layers (\texttt{mlp.fc2}, \texttt{AdaLN.modulation}, \textit{etc.}) with uniform quantizer.
		\State \Return Quantized weights, low-bit activation quantizers.
	\end{algorithmic}
\end{algorithm}
	
\section{Experiment}
\subsection{Experimental Setting}
	
\paragraph{Models and Diffusion Setup.}
Our experiments are based on the official PyTorch implementation\footnote{\url{https://github.com/facebookresearch/DiT}} of Diffusion Transformers (DiTs)~\cite{peebles2023scalable} that is pre-trained on ImageNet dataset~\cite{deng2009imagenet}.
We evaluate the proposed Hierarchical Timestep Grouping (HTG) framework using the DiT-XL/2 model for all experiments in this section.
We test our PTQ method at the image resolution of both 256×256 and 512×512. 
We employ the DDPM solver~\cite{ho2020denoising} with 100 and 50 sampling timesteps for image generation to demonstrate the generalizability of our method at diverse diffusion schedules. For all models being evaluated, we sample 10,000 images with classifier free guidance~\cite{ho2022classifier} scale=1.5, and compute the FID-10K score using the standard ADM’s TensorFlow evaluation suite~\cite{dhariwal2021diffusion}. We also report sFID~\cite{nash2021generating}, Inception Score (IS)~\cite{salimans2016improved} and Precision~\cite{kynkaanniemi2019improved} as additional metrics.

\begin{table*}[t]
	\centering
	\resizebox{0.78\textwidth}{!}{%
		\begin{tabular}{cccccccc}
			\toprule
			Timesteps             &  Bit-width~(W/A) & Method & Size~(MB) & FID $\downarrow$ & sFID $\downarrow$ & IS $\uparrow$ & Precision $\uparrow$ \\ 
			\midrule
			\multirow{17}{*}{100} & 32/32& Full Precision       & 2575.42   & 5.02 & 19.07 & 273.65 & 0.8152 \\
			\cmidrule{2-8} 
			& \multirow{7}{*}{8/8}      & PTQ4DM      & 645.72 & 15.36 & 79.31 & 172.37 & 0.6926 \\
			&                           & PTQD        & 645.72 & 8.12 & 19.64 & 199.00 & 0.7295 \\
			&                           & Q-Diffusion & 645.72 & 7.93 & 19.46 & 202.84 & 0.7299 \\
			&                           & RepQViT*     & 645.72 & 7.41 & 19.17 & 210.58 & 0.7656 \\

			&                           & PTQ4DiT*     & 645.72 & 5.42 & 20.74 & 243.32 & 0.7844 \\
			&                           &\textbf{Ours}& 652.36 & \textbf{5.14} & \textbf{20.26} & \textbf{260.86} & \textbf{0.8068} \\
			\cmidrule{2-8}
			& \multirow{7}{*}{4/8}      & PTQ4DM      & 323.79 & 89.78 & 57.20 & 26.02 & 0.2146 \\
			&                           & Q-Diffusion & 323.79 & 54.95 & 36.13 & 42.80 & 0.3846 \\
			&                           & PTQD        & 323.79 & 55.96 & 37.24 & 42.87 & 0.3948 \\
			&                           & RepQViT     & 323.79 & 26.64 & 29.42 & 91.39 & 0.4347 \\
			&                           & PTQ4DiT     & 323.79 & 7.75  & 22.01 & 190.38 & 0.7292 \\

			&                           &\textbf{Ours}& 330.43 & \textbf{6.73} & \textbf{19.61} & \textbf{219.03} & \textbf{0.7729} \\
			\midrule
			\multirow{17}{*}{50}  & 32/32  & Full Precision    & 2575.42    & 6.02 & 21.77 & 246.24 & 0.7812 \\ 
			\cmidrule{2-8} 
			& \multirow{7}{*}{8/8}      & PTQ4DM      & 645.72 & 17.52 & 84.28 & 154.08 & 0.6574 \\
			&                           & Q-Diffusion & 645.72 & 14.61 & 27.57 & 153.01 & 0.6601 \\
			&                           & PTQD        & 645.72 & 15.21 & 27.52 & 151.60 & 0.6578 \\
			&                           & RepQViT*     & 645.72 & 9.26 & 26.13 & 188.95 & 0.6980 \\

			&                           & PTQ4DiT*     & 645.72 & 6.97 & 22.13 & 214.43 & \textbf{0.7762} \\ 
			&                           &\textbf{Ours}& 648.68  & \textbf{6.78} & \textbf{21.33} & \textbf{229.09} & 0.7718  \\                    
			\cmidrule{2-8} 
			& \multirow{7}{*}{4/8}      & PTQ4DM      & 323.79 & 102.52 & 58.66 & 19.29 & 0.1710 \\
			&                           & Q-Diffusion & 323.79 & 22.89 & 29.49 & 109.22 & 0.5752 \\
			&                           & PTQD        & 323.79 & 25.62 & 29.77 & 104.28 & 0.5667 \\
			&                           & RepQViT     & 323.79 & 31.39 & 30.77 & 80.64 & 0.4091 \\
			&                           & PTQ4DiT     & 323.79 & 9.17 & 24.29 & 179.95 & 0.7052 \\  

			&                           &\textbf{Ours}& 326.75 & \textbf{8.77} & \textbf{22.31} & \textbf{189.53} & \textbf{0.7371} \\
			\bottomrule
		\end{tabular}
	}
	\vspace{-2mm}
	\caption{Performance of quantized DiT-XL/2 on ImageNet with 256$\times$256 resolution. Cfg$\_$scale$=$1.5 for all models. W/A indicates weights and activations are quantized to W bits and A bits, respectively.  Results with * are implemented using their released code. }\vspace{-3mm}
	\label{tab:exp_256}
\end{table*}

\vspace{-4mm}\paragraph{Calibration Dataset}
The calibration dataset is generated by feeding random Gaussian noise into the DiT model. We randomly generate 32 class-conditioned samples and save both the intermediate and output feature maps at each timestep for all linear layers and attention modules. No ground-truth images from the training dataset are included.

\vspace{-4mm}\paragraph{Quantization Configurations.}
We divide a diffusion process of $T$ steps into $G=\frac{T}{\lfloor 10\rfloor}$ temporal groups before calculating the corresponding shifting vector $\{\overline{\bold z}_g\}_{g=1}^G$ and scaling vector $\overline{\bold s}$ using the calibration data. We utilize the \textit{\textbf{static per-tensor}} quantizer for activations, and static per-channel quantizer for weights. The quantization parameters $\Delta$ and $\lambda$ in \cref{eq:ptq} are estimated following the wildly-used implementation from Q-Diffusion~\cite{li2023q}. We quantize the the two matrix-multiplication operations ($\bold Q\bold K^{\top}$ and $\bold A\bold V$) in self-attention module to the same precision as activation's bit-width. Softmax function remains full-precision. All quantizers use uniform quantization scheme.

\vspace{-4mm}\paragraph{Baseline Methods.}
We compare the proposed HTG method with existing post-training quantization methods. RepQViT\footnote{ \url{https://github.com/zkkli/RepQ-ViT}}~\cite{li2023repq} is design for ViT models on image classification task. PTQ4DM~\cite{shang2023ptqdm}, Q-Diffusion~\cite{li2023q} and PTQD~\cite{he2024ptqd} are three PTQ method designed for U-Net-based diffusion models. PTQ4DiT\footnote{ \url{PTQ4DiT: https://github.com/adreamwu/PTQ4DiT} }~\cite{wu2024ptq4dit} and Q-DiT~\cite{chen2024q} are specially tailor for Diffusion Transformer architecture. For the record, all the aforementioned baseline methods except Q-DiT adopt static and per-tensor activation quantizer. Therefore, we exclude Q-DiT, which utilizes dynamic activation quantization, for fair comparisons.


\subsection{Quantization Result}
We report the post-training quantization results for DiT-XL/2 models with 256$\times$256 resolution in \cref{tab:exp_256}. 
In the 100-step generation task, our method achieves state-of-the-art quantization performance at both W8A8 (8-bit weight 8-bit activation) and W4A8 precision settings in all the evaluation metrics, FID score, sFID score, IS and Precision.
The FID score of our method is 0.28 lower than that of the second best method PTQ4DiT~\cite{wu2024ptq4dit} while only needs 6.64MB more storage space in 8-bit weight setting. This storage overhead is only 1 percent of the size of the model checkpoint.
For experiments on lower bit-width precision weight, the FID score of our method is 1.02 lower than that of the second best method.
Our method also achieves state-of-the-art quantization performance at both W8A8 and W4A8 precision settings in the 50-step generation task. 

We further test the quantization performance of the proposed method on 512$\times$512 resolution DiT-XL/2 model, and report the results of W4A8 setting in \cref{tab:exp_512}. In the 100-step generation task, our method achieves state-of-the-art performance on  FID score, sFID score and Precision evaluation metrics, while the IS is slightly lower than PTQ4DiT. In the 50-step generation task, our method achieves state-of-the-art performance on all the evaluation metrics.

Sufficient experiments demonstrate that the proposed Hierarchical Timestep Grouping (HTG) method can quantize large-scale Diffusion Transformers for various scenarios with different resolutions and different denoising steps without noticeable performance degradation.

\begin{table}[t]
	\centering	\resizebox{0.99\columnwidth}{!}{%
		\begin{tabular}{cccccc}
			\toprule
			$T$            & Method     & FID $\downarrow$ & sFID $\downarrow$ & IS $\uparrow$ & Precision $\uparrow$ \\ \midrule
			\multirow{8}{*}{100} & Full Precision &  9.06 & 37.58 & 239.03 & 0.8300 \\ 
			\cmidrule{2-6} 
			& PTQ4DM     & 70.63            & 57.73             & 33.82         & 0.4574 \\
			& Q-Diffusion & 62.05            & 57.02             & 29.52         & 0.4786 \\
			& PTQD       & 81.17            & 66.58             & 35.67         & 0.5166 \\
			& RepQViT      & 62.70            & 73.29             & 31.44         & 0.3606 \\
			& PTQ4DiT      & 19.00   & 50.71    &\textbf{121.35} & 0.7514 \\
			& \textbf{Ours}     & \textbf{17.60}   & \textbf{35.09}    & 120.04 & \textbf{ 0.7679} \\
			\midrule
			\multirow{8}{*}{50}  & Full Precision        & 11.28& 41.70& 213.86 & 0.8100  \\ 
			\cmidrule{2-6} 
			& PTQ4DM     & 71.69            & 59.10             & 33.77         & 0.4604 \\
			& Q-Diffusion & 53.49            & 50.27 & 38.99 & 0.5430 \\
			& PTQD       & 73.45            & 59.14             & 39.63         & 0.5508 \\
			& RepQViT      & 65.92            & 74.19             & 30.92         & 0.3542 \\
			& PTQ4DiT      & 19.71   & 52.27             &118.32 & 0.7336 \\
			&\textbf{Ours}    & \textbf{18.95}   & \textbf{51.14}    &\textbf{118.97} & \textbf{0.7509} \\
			\bottomrule
		\end{tabular}
	}
	\label{tab:exp_512}
	\vspace{-2mm}
	\caption{Performance of quantized DiT-XL/2 on ImageNet with 512$\times$512 resolution. All models use W4/A8 quantization and Cfg$\_$scale=1.5. }\vspace{-2mm}
	
\end{table}

\subsection{Hyper-parameter Analysis}

\paragraph{Number of Temporal Groups.}
Based on the distribution similarity of activations at neighboring timesteps, we perform a hierarchical clustering algorithm to divide the diffusion process into $G$ groups in \cref{Channel-wise-shifting}. We explore the influence of taking different value of $G$ on the performance of the quantized DiT-XL/2 model. Specifically, we choose $G$ from $\{T, \lfloor\frac{T}{10}\rfloor, \lfloor\frac{T}{25}\rfloor\}$ and conduct experiments for both 100-step and 50-step diffusion processes. We also report the storage overhead required to save the corresponding bias terms $\{\widehat{\boldsymbol{\beta}}_g\}_{g=1}^G$ and $\{\ \widehat{\bold b}_g\}_{g=1}^G$. As shown in \cref{tab:G}, $G=\lfloor\frac{T}{10}\rfloor$ not only achieves almost the same performance compared with dividing each timestep into a separate group, but also $G=\lfloor\frac{T}{10}\rfloor$ significantly reduces the storage overhead after quantization, considering that the 4-bit DiT-XL/2 model is $\sim$324MB and the 8-bit model is $\sim$653MB.

\begin{table}[t]
\small
\centering
	\begin{tabular}{ccccc}
		\toprule
		Timesteps & $G$ & Storage Overhead & FID $\downarrow$ & sFID $\downarrow$ \\
		\midrule
		\multirow{3}{*}{$T=100$}&100& 73.09 MB& \textbf{6.70} & \textbf{19.52} \\
		& 10 & 6.64 MB&  6.73 & 19.61 \\
		& 4  & 2.21 MB& 6.83 & 19.85 \\
		\midrule
		\multirow{3}{*}{$T=50$} & 50 & 36.17 MB&8.81 & \textbf{22.14} \\
		& 5 & 2.96 MB& \textbf{8.77} & 22.31 \\
		& 2 & 1.48 MB& 9.23 & 26.16\\
		\bottomrule
	\end{tabular}
	\vspace{-2mm}
	\caption{The impact of different numbers of groups on W4/A8 quantization performance. }\vspace{-5mm}
	\label{tab:G}
\end{table}

\vspace{-4mm}\paragraph{EMA Coefficient.}
\cref{eq:ema} provides a way to aggregate the historical per-channel outlier information by take the exponential moving average (EMA) of the activation's maximum value. We find that the final performance of the quantized DiT model is influenced by the value of EMA coefficient $\alpha$.
We test with different value of $\alpha$ from $\{0.9, 0.99, 0.999\}$ by quantizing a DiT-XL/2 model into W4A8 precision with a 100-step diffusion process, and report the results in \cref{tab:ema_coeff}.
$\alpha=0.99$ leads to the best result.

\subsection{Ablation Study on Each Operation}
To validate the effectiveness of the proposed Temporally-grouped Channel-wise Shifting (\cref{Channel-wise-shifting}) and Temporally-aggregated Channel-wise Scaling (\cref{Channel-wise-scaling}) operations, we conduct an in-depth ablation study in which we quantize the DiT-XL/2 into W4A8 precision by different means of quantization schemes. We employ the 100-step DDPM solver and 256$\times$256 image resolution. The fundamental uniform quantization are used as our baseline method. We then gradually incorporate Temporally-grouped Channel-wise Shifting (denoted by \textbf{op1}) and Temporally-aggregated Channel-wise Scaling (denoted by \textbf{op2}). The performance of the quantized models are shown in \cref{tab:ablation}. 
As the table suggests, both operations "Temporally-grouped Channel-wise Shifting" and "Temporally-aggregated Channel-wise Scaling" improve the performance compared to the baseline, with their combination (the proposed HTG framework) yielding the best result. In spite of the fact that the scaling operation plays a more important role, using the channel-wise scaling (\textbf{op2}) alone could not reach an ideal quantization result, since the temporally varying outliers cannot be fully smoothed without the per-channel shifting operation (\textbf{op1}) to help re-adjusting and zero-centering the asymmetric outlier distribution.

\begin{table}[t]
	\small
	\centering
	\resizebox{0.41\columnwidth}{!}{%
		\begin{tabular}{ccc}
			\toprule
			$\alpha$ & FID $\downarrow$ & sFID $\downarrow$ \\
			\midrule
			0.9 &  7.21 & 20.59 \\
			0.99 &  \textbf{6.73} & \textbf{19.61} \\
			0.999 &  6.75 & 19.63 \\
			\bottomrule
		\end{tabular}
	}
	\vspace{-2mm}
	\caption{Performance with different EMA coefficients. }\vspace{-2mm}
	\label{tab:ema_coeff}
\end{table}

\begin{table}[t]
	\small
	\centering
		\begin{tabular}{ccc}
			\toprule
			Method & FID $\downarrow$ & sFID $\downarrow$ \\
			\midrule
			Full Precision &  5.02 & 19.07 \\
			Baseline &  29.65 &  44.92 \\
			Baseline + \textbf{op1} & 18.90 & 27.13 \\
			Baseline + \textbf{op2} & 7.52 & 21.93 \\
			Baseline + \textbf{op1} + \textbf{op2}  &  6.73 & 19.61 \\
			\bottomrule
		\end{tabular}
		\vspace{-2mm}
		\caption{Ablation study on the effect of each operation. }\vspace{-4mm}
	\label{tab:ablation}
\end{table}
	
	\vspace{-1mm}
\section{Conclusion}\vspace{-1mm}
In this work, we first identify the unique challenges of post-training quantization for Diffusion.
To overcome these challenges, we introduce the Hierarchical Timestep Grouping (HTG) framework, a novel quantization approach designed for DiT models. HTG incorporates temporally-grouped channel-wise shifting and temporally-aggregated channel-wise scaling operations to dynamically adjust the activation distributions across different timesteps. This ensures that the activations are quantized efficiently, with minimal error accumulation during the iterative denoising process. We further utilize a clustering-based approach to reduce the storage overhead, and a re-parameterization technique to avoid introducing additional computations.
We conduct extensive experiments to validate the effectiveness of the HTG framework, which demonstrate significant improvements in quantization performance for large-scale DiT models over existing post-training quantization baselines.

\newpage
{
    \small
    \bibliographystyle{ieeenat_fullname}
    \bibliography{main}
}

\end{document}